\documentclass{article}

\usepackage{times}
\usepackage{graphicx} 
\usepackage{subcaption}

\usepackage[cmex10]{amsmath}
\usepackage{amsfonts}
\usepackage[noend]{algorithmic}
\usepackage{array}
\usepackage{multirow}
\usepackage{caption}
\usepackage{verbatim}
\usepackage{perpage}

\usepackage{amsthm}
\usepackage{graphicx}
\usepackage{booktabs}
\usepackage{lineno}
\usepackage{natbib}
\usepackage{amssymb}

\usepackage{algorithm}
\usepackage{hyperref}

\usepackage[accepted]{whi2016} 

\MakePerPage{footnote} 

\newcommand{\argmin}{\operatornamewithlimits{argmin}}

\icmltitlerunning{Interpretability in Brain Decoding}

\newtheorem{mydefs}{Definition}
\newtheorem*{myprobs}{Problem}

\begin{document} 
\twocolumn[
\icmltitle{Interpretability in Linear Brain Decoding}

\icmlauthor{Seyed Mostafa Kia}{seyedmostafa.kia@unitn.it}
\icmlauthor{Andrea Passerini}{andrea.passerini@unitn.it}
\icmladdress{Department of Information Engineering and Computer Science, University of Trento, Via Sommarive, 9 I-38123 Povo, Trento, Italy}

\icmlkeywords{machine learning, brain decoding, interpretation, model selection}

\vskip 0.3in
]

\begin{abstract} 
Improving the interpretability of brain decoding approaches is of primary interest in many neuroimaging studies. Despite extensive studies of this type, at present, there is no formal definition for interpretability of brain decoding models. As a consequence, there is no quantitative measure for evaluating the interpretability of different brain decoding methods. In this paper, we present a simple definition for interpretability of linear brain decoding models. Then, we propose to combine the interpretability and the performance of the brain decoding into a new multi-objective criterion for model selection. Our preliminary results on the toy data show that optimizing the hyper-parameters of the regularized linear classifier based on the proposed criterion results in more informative linear models. The presented definition provides the theoretical background for quantitative evaluation of interpretability in linear brain decoding.
\end{abstract} 

\section{Introduction}
\label{sec:introduction}
In cognitive science, researchers usually analyze recorded brain activity to discover the answers of \emph{where}, \emph{when}, and \emph{how} a brain region participates in a particular cognitive process. To answer the key questions in cognitive science, scientists often employ mass-univariate hypothesis testing methods to test scientific hypotheses on a large set of independent variables~\cite{groppe2011amass}. On the down side, the high dimensionality of neuroimaging data requires a large number of tests that reduces the sensitivity of these methods after multiple comparison correction. The multivariate counterparts of mass-univariate analysis, known generally as multivariate pattern analysis (MVPA), have the potential to overcome this deficit.

\emph{Brain decoding}~\cite{haynes2006decoding} is an MVPA technique that delivers a model to predict the mental state of a human subject based on the recorded brain signal. From the neuroscientific perspective, a brain map resulting from weight of linear brain decoding model is considered \emph{interpretable} if it enables the scientist to answer \emph{where}, \emph{when}, and \emph{how} questions. But typically a classifier, taken alone, only answers the question of \emph{what} is the most likely label of a given unseen sample. This fact is generally known as knowledge extraction gap~\cite{vellido2012making} in the classification context. Thus far, many efforts have been devoted to filling the knowledge extraction gap of linear and non-linear data modeling methods in different areas such as computer vision~\cite{bach2015pixel}, signal processing~\cite{montavon2013analyzing}, chemometrics~\cite{yu2015classification}, bioinformatics~\cite{hansen2011visual}, and neuroinformatics~\cite{haufe2013interpretation}.

Despite the theoretical advantages of MVPA, its practical application to inferences regarding neuroimaging data is limited primarily due to the knowledge extraction gap~\cite{sabuncu2014universal}. Therefore, improving the interpretability of linear brain decoding and associated brain maps is a primary goal in the brain imaging literature~\cite{strother2014stability}. The lack of interpretability of multivariate brain maps is a direct consequence of low signal-to-noise ratios (SNRs), high dimensionality of whole-scalp recordings, high correlations among different dimensions of data, and cross-subject variability. At present, two main approaches are proposed to enhance the interpretability of multivariate brain maps: 1) introducing new metrics, such as reproducibility of maps or stability of models, into the model selection procedure~\cite{rasmussen2012model,conroy2013fast,yu2013stability}, and 2) introducing new hybrid penalty terms for regularization to incorporate spatio-temporal prior knowledge in the learning~\cite{van2009interpreting,michel2011total,de2012combining,grosenick2013interpretable}.

In spite of the aforementioned efforts to improve the interpretability, there is still no formal definition for the interpretability of brain decoding in the literature. Therefore, the interpretability of different brain decoding methods are evaluated either qualitatively or indirectly. With the aim of filling this gap, our contribution is two-fold: 1) assuming that the true solution of brain decoding is available, we present a simple definition of the interpretability in linear brain decoding; 2) we propose the combination of the interpretability and the performance of the brain decoding as a new Pareto optimal multi-objective criterion for model selection. We experimentally, on a toy dataset, show that incorporating the interpretability into the model selection procedure provides more interpretable models~\footnote{\tiny For further experiments on the real dataset please see~\cite{kia2016interpretability}}.
\section{Methods}
\label{sec:methods}
\subsection{Notation and Background}
\label{subsec:notations}
Let $\mathcal{X} \in \mathbb{R}^p$ be a manifold in Euclidean space that represents the input space and $\mathcal{Y}\in \mathbb{R}$ be the output space, where $\mathcal{Y}=\Phi^*(\mathcal{X})$. Then, let $S=\{ \textbf{Z} = (\textbf{X},\textbf{Y}) \mid z_1=(x_1,y_1), \dots , z_n=(x_n,y_n)\}$ be a training set of $n$ independently and identically distributed (iid) samples drawn from the joint distribution of $\mathcal{Z} = \mathcal{X} \times \mathcal{Y}$. In the neuroimaging context, $\textbf{X}$ indicates the trials of brain recording, and $\textbf{Y}$ represents the experimental conditions. The goal of brain decoding is to find the function $\Phi_S: \textbf{X} \to \textbf{Y}$ as an estimation of the ideal function $\Phi^*: \mathcal{X} \to \mathcal{Y}$.

As is a common assumption in the neuroimaging context, we assume the true solution of a brain decoding problem is among the family of linear functions $\mathcal{H}$. Therefore, the aim of brain decoding reduces to finding an empirical approximation of $\Phi_S$, indicated by $\hat\Phi$, among all $\Phi \in \mathcal{H}$. This approximation can be obtained by solving a risk minimization problem:
\begin{eqnarray} \label{eq:max2min}
\begin{split}
\hat{\Theta} = \argmin_{\Theta} \mathcal{L}(\textbf{Y},\Phi_S(\textbf{X}))+ \lambda \Omega (\Theta)
\end{split}
\end{eqnarray}

where $\Theta$ denotes the parameters of the linear model, $\mathcal{L}: \textbf{Z} \times \textbf{Z} \to \mathbb{R}^+$ is the loss function, $\Omega:\mathbb{R}^{p}\to \mathbb{R}^+$ is the regularization term, and $\lambda$ is a hyper-parameter that controls the amount of regularization. $\lambda$ is generally decided using cross-validation or other data perturbation methods in the model selection procedure.

The estimated parameters of a linear decoding model $\hat{\Theta}$ can be used in the form of a brain map so as to visualize the discriminative neurophysiological effect. We refer to the normalized parameter vector of a linear brain decoder in the unit hyper-sphere as a multivariate brain map (MBM); we denote it by $\vec{\Theta}$ where $\vec{\Theta} = \frac{\Theta}{\left \| \Theta \right \|}$ ($\left \| . \right \|$ is the 2-norm).

As shown in Eq.~\ref{eq:max2min}, learning occurs using the sampled data. In other words, in the learning paradigm, we attempt to minimize the loss function with respect to $\Phi_S$ (and not $\Phi^*$)~\cite{poggio2002mathematical}. The \emph{irreducible error} $\varepsilon$ is the direct consequence of sampling; it sets a lower bound on the error, where we have:
\begin{eqnarray} \label{eq:estimation}
\Phi_S(\textbf{X}) = \Phi^*(\textbf{X}) + \varepsilon
\end{eqnarray}

\subsection{Theoretical Definition}
\label{subsec:interpretability_MBM}
In this section, we present a definition for the interpretability of linear brain decoding models and their associated MBMs. Our definition of interpretability is based on two main assumptions: 1) the brain decoding problem is linearly separable; 2) its \emph{unique} and neurophysiologically \emph{plausible} solution, i.e., $\Phi^*$, is available. 

Consider a linearly separable brain decoding problem in an ideal scenario where $\varepsilon=0$ and $rank(\textbf{X})=p$. In this case, $\Phi^*$ is linear and its parameters $\Theta^*$ are unique and plausible. The unique parameter vector $\Theta^*$ can be computed by:
\begin{eqnarray} \label{eq:ideal_least_square}
\Theta^* = \Sigma_{\textbf{X}}^{-1} \textbf{X}^T \textbf{Y}
\end{eqnarray}

$\Sigma_{\textbf{X}}$ represents the covariance of $\textbf{X}$. Using $\Theta^*$ as the reference, we can define the \emph{strong-interpretability}:
\begin{mydefs} \label{def:strong_interpretability}
An MBM $\vec\Theta$ associated with a linear function $\Phi$ is ``strongly-interpretable" if and only if $\vec\Theta \propto \Theta^*$.
\end{mydefs}

In practice, the estimated solution of a linear brain problem is not strongly-interpretable because of the inherent limitations of neuroimaging data, such as uncertainty~\cite{aggarwal2009survey} in the input and output space ($\varepsilon \neq 0$), the high dimensionality of data ($n \ll p$), and the high correlation between predictors ($rank(\textbf{X})<p$). With these limitations in mind, even though the solution of linear brain decoding is not strongly-interpretable, one can argue that some are more interpretable than others. For example, in the case in which $\Theta^* \propto [0,1]^T$, a linear classifier where $\hat{\Theta} \propto [0.1,1.2]^T$ can be considered more interpretable than a linear classifier where $\hat{\Theta} \propto [2,1]^T$. This issue raises the following question: 
\begin{myprobs} \label{prob:interpretability}
Let $S^1, \dots, S^m$ be $m$ perturbed training sets drawn from $S$ via a certain perturbation scheme such as bootstrapping, or cross-validation. Assume $\vec{\hat\Theta}^1, \dots, \vec{\hat\Theta}^m$ are $m$ MBMs of a certain $\Phi$ on the corresponding perturbed training sets. How can we quantify the proximity of $\Phi$ to the strongly-intrepretable solution of brain decoding problem $\Phi^*$?  
\end{myprobs}

Considering the uniqueness and the plausibility of $\Phi^*$ as the two main characteristics that convey its strong-interpretability, we define the interpretability as follows:
\begin{mydefs} \label{def:interpretability}
Let $\alpha^j$ ($j = 1,\dots,m$) be the angle between $\vec{\hat\Theta}^j$ and $\vec{\Theta}^*$. The ``interpretability" ($0 \leq \eta_\Phi \leq 1$) of the MBM derived from a linear function $\Phi$ is defined as:
\begin{eqnarray} \label{eq:interpretability}
\eta_\Phi = \frac{1}{m} \sum_{j=1}^m \cos(\alpha^j)
\end{eqnarray}
\end{mydefs}

In fact, the interpretability is the average cosine similarities between $\Theta^*$ and MBMs derived from different samplings of the training set. Even though, in practice, the exact computation of $\eta_\Phi$ is unrealistic (as $\Theta^*$ is not available), the interpretability of the decoding model can be approximated based on ad-hoc heuristics (see~\cite{kia2016interpretability} for an example in the magnetoenecephalography decoding). The approximated interpretability can be incorporated in the model selection procedure in order to find more reproducible and plausible decoding models. 

\subsection{Interpretability in Model Selection}
\label{subsec:interpretability_model_selection}
The procedure for evaluating the performance of a model so as to choose the best values for hyper-parameters is known as \emph{model selection}~\cite{hastie2009elements}. This procedure generally involves numerical optimization of the model selection criterion. The most common model selection criterion is based on an estimator of generalization performance. In the context of brain decoding, especially when the interpretability of brain maps matters, employing the predictive power as the only decisive criterion in model selection is problematic~\cite{rasmussen2012model,conroy2013fast}. Here, we propose a multi-objective criterion for model selection that takes into account both prediction accuracy and MBM interpretability.

Let $\eta_\Phi$ and $\delta_\Phi$ be the interpretability and the generalization performance of a linear function $\Phi$, respectively. We propose the use of the \emph{scalarization} technique~\cite{caramia2008multi} for combining $\eta_\Phi$ and $\delta_\Phi$ into one scalar $0 \leq \zeta(\Phi) \leq 1$ as follows:
\begin{eqnarray} \label{eq:plausibility}
\zeta_\Phi =
\left\{\begin{matrix}
\frac{\omega_1 \eta_\Phi + \omega_2 \delta_\Phi}{\omega_1 + \omega_2}  & \delta_\Phi \geq \kappa \\
0 & \delta_\Phi < \kappa
\end{matrix}\right.
\end{eqnarray}

where $\omega_1$ and $\omega_2$ are weights that specify the importance of the interpretability and the performance, respectively. $\kappa$ is a threshold that filters out solutions with poor performances. In classification scenarios, $\kappa$ can be set by adding a small safe interval to the chance level. It can be shown that the hyper-parameters of a model $\Phi$ are optimized based on $\zeta_\Phi$ are Pareto optimal~\cite{marler2004survey}.

\subsection{Classification and Evaluation}
\label{subsubsec:classification_evaluation}
In our experiment, a least squares classifier with L1-penalization, i.e., Lasso~\cite{tibshirani1996regression}, is used for decoding. Lasso is a popular classification method in brain decoding, mainly because of its sparsity assumption. The choice of Lasso helps us to better illustrate the importance of including the interpretability in the model selection. Lasso solves the following optimization problem:
\begin{eqnarray} \label{eq:lasso}
\hat{\Theta} = \argmin_{\Theta} \left\| \Phi(\textbf{X})-\Phi_S(\textbf{X}) \right\|_2^2 + \lambda \left \| \Theta \right\|_1
\end{eqnarray}

where $\lambda$ is the hyper-parameter that specifies the level of regularization. Therefore, the aim of the model selection is to find the best value for $\lambda$. Here, we try to find the best regularization parameter value among $\lambda=\{0.001, 0.01, 0.1, 1, 10, 50, 100, 250, 500, 1000\}$.

We use the out-of-bag (OOB)~\cite{breiman2001random} method to compute $\delta_\Phi$, $\eta_\Phi$, and $\zeta_\Phi$ for different values of $\lambda$. In OOB, given a training set $(\textbf{X},\textbf{Y})$, $m$ replications of bootstrap are used to create perturbed training sets (we set $m = 50$)~\footnote{\tiny The MATLAB code used for experiments is available at \url{https://github.com/smkia/interpretability/}}. We set $\omega_1 = \omega_2 = 1$ and $\kappa=0.6$ in the computation of $\zeta_\Phi$. Furthermore, we set $\delta_\Phi=1-EPE$ where EPE indicates the expected prediction error.
\section{Experiment}
\label{sec:experiment}
\subsection{Experimental Material}
\label{subsec:materials}
\begin{figure}
         \centering
         \begin{subfigure}[h]{0.45\textwidth}
         	\includegraphics[width=\textwidth]{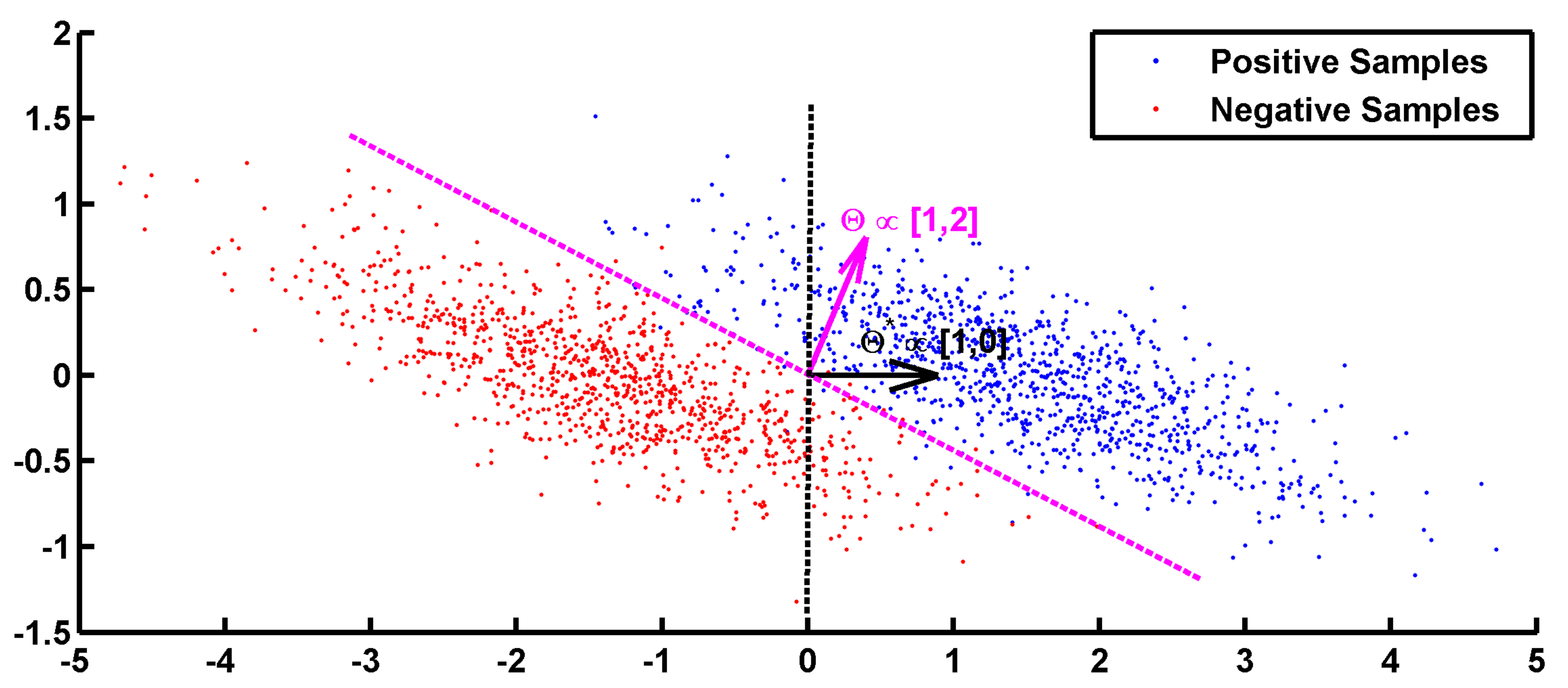}
         \end{subfigure}
         \caption{Noisy samples of toy data. The black line shows the true separator based on the generative model ($\Phi^*$). The magenta line shows the most accurate classification solution.}
         \label{fig:simulation_study_1_data}
\end{figure}

To illustrate the importance of integrating the interpretability of brain decoding with the model selection procedure, we use simple 2-dimensional toy data presented in~\cite{haufe2013interpretation}. Assume that the true underlying generative function $\Phi^*$ is defined by:
\begin{align*}
\mathcal{Y}=\Phi^*(\mathcal{X})=\left\{\begin{matrix}
1 & \quad if \quad \textit{x}_1 = 1.5 \\
-1 & \quad if \quad \textit{x}_1 = -1.5
\end{matrix}\right.
\end{align*}
where $\mathcal{X} \in \{ [1.5,0]^T, [-1.5,0]^T\}$; and $x_1$ and $x_2$ represent the first and the second dimension of the data, respectively. Furthermore, assume the data is contaminated by Gaussian noise with co-variance $\Sigma=\begin{bmatrix} 1.02 & -0.3\\ -0.3 & 0.15 \end{bmatrix}$. Figure~\ref{fig:simulation_study_1_data} shows the distribution of the noisy data.

\subsection{Results}
\label{subsec:results}
\begin{table*}[]
\centering
\caption{Comparison between $\delta_\Phi$, $\eta_\Phi$, and $\zeta_\Phi$ for different $\lambda$ values on the toy 2D example shows the performance-interpretability dilemma, in which the most accurate classifier is not the most interpretable one.}
\label{tab:simulation_study_1_results}
\tiny\setlength{\tabcolsep}{2.5pt}
\begin{tabular}{@{}l|ccccccccccc@{}}
\toprule
$\lambda$                    & 0                                                & 0.001                                             & 0.01                                             & 0.1                                              & 1                                                & 10                                               & 50                                               & 100                                             & 250                                   & 500                                   & 1000                                  \\ \midrule
$\delta(\Phi)$                          & 0.9883                                           & 0.9883                                            & 0.9883                                           & 0.9883                                           & 0.9883                                           & \textbf{0.9884}                                  & 0.9880                                           & 0.9840                                          & 0.9310                                & 0.9292                                & 0.9292                                \\
$\eta(\Phi)$           & 0.4391                                           & 0.4391                                            & 0.4391                                           & 0.4392                                           & 0.4400                                           & 0.4484                                           & 0.4921                                           & 0.5845                                          & 0.9968                                & \textbf{1}                            & \textbf{1}                            \\
$\zeta(\Phi)$          & 0.7137                                           & 0.7137                                            & 0.7137                                           & 0.7137                                           & 0.7142                                           & 0.7184                                           & 0.7400                                           & 0.7842                                          & 0.9639                                & \textbf{0.9646}                       & \textbf{0.9646}                       \\
$\vec{\hat{\Theta}} \propto$ & $\begin{bmatrix} 0.4520 \\ 0.8920 \end{bmatrix}$ & $\begin{bmatrix} 0.4520 \\ 0.8920  \end{bmatrix}$ & $\begin{bmatrix} 0.4520 \\ 0.8920 \end{bmatrix}$ & $\begin{bmatrix} 0.4521 \\ 0.8919 \end{bmatrix}$ & $\begin{bmatrix} 0.4532 \\ 0.8914 \end{bmatrix}$ & $\begin{bmatrix} 0.4636 \\ 0.8660 \end{bmatrix}$ & $\begin{bmatrix} 0.4883 \\ 0.8727 \end{bmatrix}$ & $\begin{bmatrix}0.5800 \\ 0.8146 \end{bmatrix}$ & $\begin{bmatrix}0.99 \\ 0.02 \end{bmatrix}$ & $\begin{bmatrix}1 \\ 0 \end{bmatrix}$ & $\begin{bmatrix}1 \\ 0 \end{bmatrix}$ \\ \bottomrule
\end{tabular}
\end{table*}

In the definition of $\Phi^*$ on the toy dataset, $x_1$ is the decisive variable and $x_2$ has no effect on the classification of the data into target classes. Therefore, excluding the effect of noise and based on the theory of the maximal margin classifier, $\vec{\Theta}^* \propto [1,0]^T$ is the true solution to the decoding problem. By accounting for the effect of noise and solving the decoding problem in $(\textbf{X},\textbf{Y})$ space, we have $\vec{\Theta} \propto [\frac{1}{\sqrt(5)},\frac{2}{\sqrt(5)}]^T$ as the parameter of the linear classifier. Although the estimated parameters on the noisy data yield the best generalization performance for the noisy samples, any attempt to interpret this solution fails, as it yields the wrong conclusion with respect to the ground truth (it says $x_2$ has twice the influence of $x_1$ on the results, whereas it has no effect). This simple experiment shows that the most accurate model is not always the most interpretable one, primarily because the contribution of the noise in the decoding process~\cite{haufe2013interpretation}. On the other hand, the true solution of the problem $\vec{\Theta}^*$ does not provide the best generalization performance for the noisy data.

To illustrate the effect of incorporating the interpretability in the model selection, a Lasso model with different $\lambda$ values is used for classifying the toy data. In this case, because $\vec{\Theta}^*$ is known, the interpretability can be computed using Eq.~\ref{eq:interpretability}. Table~\ref{tab:simulation_study_1_results} compares the resultant performance and interpretability from Lasso. Lasso achieves its highest performance ($\delta_\Phi = 0.9884$) at $\lambda=10$ with $\vec{\hat{\Theta}} \propto [0.4636, 0.8660]^T$ (indicated by the magenta line in Figure~\ref{fig:simulation_study_1_data}). Despite having the highest performance, this solution suffers from a lack of interpretability ($\eta_\Phi=0.4484$). By increasing $\lambda$, the interpretability improves so that for $\lambda = 500, 1000$ the classifier reaches its highest interpretability by compensating for $0.06$ of its performance. Our observation highlights two main points: 1) In the case of noisy data, the interpretability of a decoding model is incoherent with its performance. Thus, optimizing the parameter of the model based on its performance does not necessarily improve its interpretability. This observation confirms the previous finding by~\citet{rasmussen2012model} regarding the trade-off between the spatial reproducibility (as a measure for the interpretability) and the prediction accuracy in brain decoding; 2) if the right criterion is used in the model selection, employing proper regularization technique (sparsity prior, in this case) leads to more interpretability for the decoding models.
\section{Discussions}
\label{sec:discussions}
In this study, our primary interest was to present a definition of the interpretability of linear brain decoding models. Our definition and quantification of interpretability remains theoretical, as we assume that the true solution of the brain decoding problem is available. Despite this limitation, we argue that the presented simple definition provides a concrete framework of a previously abstract concept and that it establishes a theoretical background to explain an ambiguous phenomenon in the brain decoding context. 

Despite ubiquitous use, the generalization performance of classifiers is not a reliable criterion for assessing the interpretability of brain decoding models~\cite{rasmussen2012model}. Therefore, considering extra criteria might be required. However, because of the lack of a formal definition for interpretability, different characteristics of brain decoding models are considered as the main objective in improving their interpretability. Our definition of interpretability helped us to fill this gap by introducing a new multi-objective criterion as a weighted compromise between interpretability and generalization performance. Furthermore, this work presents an effective approach for evaluating the quality of different regularization strategies for improving the interpretability of MBMs. Our findings provide a further step toward direct evaluation of interpretability of the currently proposed penalization strategies. 

Despite theoretical advantages, the proposed definition of interpretability suffer from some limitations. The presented concepts are defined for linear models, with the main assumption that $\Phi^* \in \mathcal{H}$ (where $\mathcal{H}$ is a class of linear functions). Extending the definition of interpretability to non-linear models demands future research in visualization of non-linear models in the form of brain maps.

\bibliography{interpretationbib}
\bibliographystyle{icml2016}

\end{document}